# PANSHARPENING OF PRISMA PRODUCTS FOR ARCHAEOLOGICAL PROSPECTION

*Gregory Sech\*, Giulio Poggi\*, Marina Ljubenović, Marco Fiorucci, Arianna Traviglia*

Center for Cultural Heritage Technology, Istituto Italiano di Tecnologia, Venice, Italy

## ABSTRACT

Hyperspectral data recorded from satellite platforms are often ill-suited for geo-archaeological prospection due to low spatial resolution. The established potential of hyperspectral data from airborne sensors in identifying archaeological features has, on the other side, generated increased interest in enhancing hyperspectral data to achieve higher spatial resolution. This improvement is crucial for detecting traces linked to sub-surface geo-archaeological features and can make satellite hyperspectral acquisitions more suitable for archaeological research.

This research assesses the usability of pansharpened PRISMA satellite products in geo-archaical prospections. Three pan-sharpening methods (GSA, MTF-GLP and HySure) are compared quantitatively and qualitatively and tested over the archaeological landscape of Aquileia (Italy). The results suggest that the application of pansharpening techniques makes hyperspectral satellite imagery highly suitable, under certain conditions, to the identification of sub-surface archaeological features of small and large size.

## 1. INTRODUCTION

Despite their limited availability hindering widespread use, hyperspectral data have found application in archaeological research due to their ability to enhance the visibility of surface traces and anomalies related to the presence of sub-soil archaeological deposits [1], [2], [3]. The majority of archaeological applications have predominantly utilised data from airborne sensors, with satellite hyperspectral data representing a relatively novel frontier in this field. Due to the inherent trade-off between spatial and spectral resolution in the design of hyperspectral sensors mounted on satellites, the efficiency of satellite acquisitions is compromised when it comes to detecting small-scale features, which are often prevalent in the archaeological and geomorphological domains. Pansharpening techniques can effectively be used to enhance the spatial resolution of multispectral and hyperspectral products [4].

PRISMA, the Earth observation system launched by the Italian Space Agency (ASI) in 2019, represents a promising opportunity for archaeological remote sensing research. The hyperspectral sensor mounted on the satellite provides 240 bands in the VNIR-SWIR range at 30m Ground Sampling Distance (GSD), coupled with a panchromatic acquisition at 5m GSD. The short revisit period of 7 days at variable look angles enables frequent and comprehensive coverage, making PRISMA data well-suited for time-series analysis. ASI makes available PRISMA products free of charge under an open licence for registered research projects, promoting the widespread utilisation of hyperspectral data on a global scale. To our knowledge, the use of PRISMA data in archaeological research is very limited thus far [5]: research undertaken underscores the potential of PRISMA's spectral capabilities, albeit somewhat constrained by its low spatial resolution.

This paper aims to offer the first comparative evaluation of pansharpening techniques applied to PRISMA hyperspectral data in the context of geo-archaeological prospection. Drawing from the findings of the 2022 WHISPERS Hyperspectral Pansharpening Challenge [6], a comparative analysis is undertaken between the two top-performing baseline techniques, namely GLS and MTF-GLP, and the HySure [7] method. A qualitative commentary on the results of the pansharpening techniques is provided, taking into account the perspective of the geo-archaeological practice. The discussion explores the practical relevance of the quantitative metrics commonly used in current evaluation of pansharpening methods.

## 2. CASE STUDY

Pansharpened products from PRISMA were assessed by comparing images depicting the region surrounding Aquileia, formerly a Roman city established in the 2$^{nd}$ century BCE in northeastern Italy (Fig.1). The area encompasses a portion of an extensive coastal plain where buried infrastructures (roads and land divisions), settlements, remnants of palaeo-hydrographic networks (palaeochannels) and lagoonal areas are intricately interconnected: these elements form integral

---

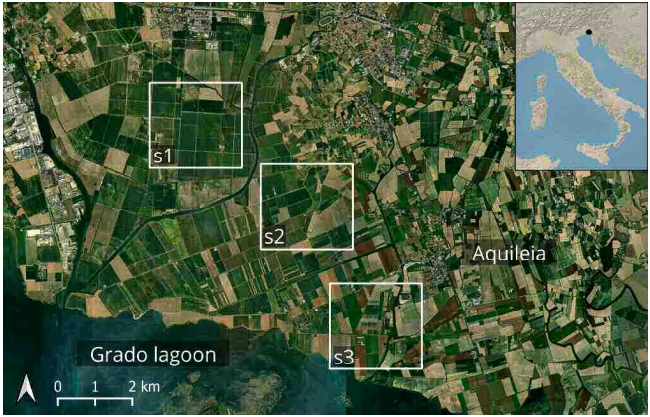

**Figure 1 - Aquileia and its territory. Sectors 1-3 indicate focus areas used for visual and qualitative assessment.**

components of the ancient landscape development. The combination of agricultural practices with a flat topography enhances the visibility of traces and anomalies associated with subsoil features on low vegetation and bare soil.
Two PRISMA products capturing the same area (acquired on June 4, 2022 and August 24, 2023) were chosen as test images based on catalogue availability and minimal cloud coverage. Three main sub-samples (sector s1, s2, s3) were then selected from the entire area for visual inspection and qualitative analysis.

### 3. METHODOLOGY

**3.1. Pre-processing of PRISMA products**

The PRISMA hyperspectral and panchromatic products have been successfully co-registered with a Sentinel-2 reference image using AROSICS software [8], following a well-established procedure for dealing with poorly co-registered data [6]. Subsequently, atmospheric absorption bands and bands with low signal-to-noise (SNR) ratio were removed following the procedure outlined in [9]. The remaining bands for analysis numbered 167. To apply pansharpening across all the bands, the rasters were divided in tiles and eventually padded to 360x360 pixels at 5m GSD. After applying each pansharpening method, the resulting tiles were unpadded and merged back into the original raster size (7600x7400 pixels). This facilitated both visual inspection and the computation of quantitative metrics.

**3.2. Pansharpening Methods**

The pansharpening method HySure [7] has been compared to the two top-performing models emerging from the WHISPER challenge: GSA [10] and MTF-GLP [11]. HySure models the physical relationship between the spectral response of the panchromatic band and the hyperspectral data cube. Additionally, it demonstrates robustness to potential sensor's deviation from factory specifications, as it estimates sensor properties solely from the acquired data.

The Gram-Schmidt Adaptive (GSA) method represents an improvement over the well-known Gram-Schmidt component substitution pansharpening approach [10]. This improvement results from the application of a linear regression to minimise the error between the degraded panchromatic observation and the up-sampled hyperspectral product. The weights obtained through the regression are subsequently used in the inverse of the Gram-Schmidt transform.

In the Modulation Transfer Function - Generalised Laplacian Pyramid (MTF-GLP) pansharpening method, a hyperspectral and a panchromatic image are merged using multi-resolution analysis (MRA) [11]. The method uses a sensor-tailored generalised Laplacian pyramid to find high-pass details in a high-resolution panchromatic band that have been dampened by the hyperspectral optical sensor's Modulation Transfer Function (MTF).

The HySure method defines data-fusion as an inversion problem using $Y_h$ to model hyperspectral measurements and $Y_m$ for the multispectral (in our case panchromatic) measurements. The formulated optimisation problem is:

$$\min_X \frac{1}{2} \|Y_h - EXBM\|_F^2 + \frac{\lambda_m}{2}\|Y_m - REX\|_F^2 + \lambda_\varphi \varphi(XD_h, XD_v),$$

where the regulariser $\varphi$ is Vector Total Variation (VTV) with the regularisation parameter $\lambda_\varphi$. R is a row vector containing the spectral response of the panchromatic band. M is a uniform sub-sampling operation and B is a spatial blurring matrix representing the hyperspectral sensor's Point Spread Function (PSF) in the spatial resolution of $Y_m$. The PSF is closely related to the sensor's MTF used in the MTF-GLP; however, it is estimated directly from data [12]. Matrix E models the hypothesis that hyperspectral data lies in a lower-dimensional manifold and is computed in our implementation using Vertex Component Analysis (VCA) [13]. X stands for the subspace representation coefficients. The optimisation problem is convex and a global minima can be computed efficiently using the SALSA algorithm [14].

**3.3 Quantitative assessment**

Evaluation of the properties of pansharpened products relies on two settings: Wald's protocol and a full-resolution evaluation. Wald's protocol employs metrics such as UIQI, SAM, and ERGAS to evaluate the method's effectiveness in reconstructing the original product. This is accomplished by comparing the hyperspectral image with the image obtained by applying the pansharpening technique to a down-sampled version of the same PRISMA product.

The evaluation of the data-fusion result at full-resolution measures spatial consistency with panchromatic data using metric $D_s^*$ [15]. Spectral distortion with the hyperspectral cube is measured using $D_\lambda^k$. Lastly the $Q^*$ index is computed to quantify spatial and spectral distortions in a single metric.

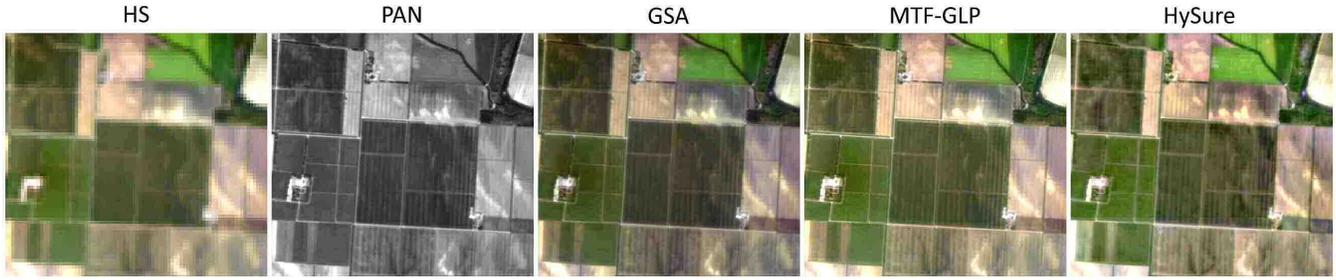

**Figure 2 – Sector s1. Comparison of PRISMA products (HS and PAN) with the results of the pansharpening methods.**

## 3.4 Qualitative assessment

Visual inspection was performed to evaluate the effectiveness of the pansharpening methods in enhancing archaeological and geomorphological features. Pansharpened images were further processed to exploit the spectral capabilities of all 66 bands of the VNIR cube (400 – 1010 nm) that can be used to evaluate the vegetation status, and therefore sensitive to alterations caused by sub-surface features [16]. Dimensionality reduction with Principal Components Analysis (PCA) was applied to condense information contained in the large spectrum into a limited number of bands, thereby drastically reducing noise and artefacts resulted from the pansharpening. Based on the PCA results, false-colour composites images were created by selecting the three most informative components (PCs) following an initial visual evaluation of the outputs. The resulting images were then visually compared with a ground-truth dataset collected in the last 15 years of studies, incorporating data from various types of sensors and spatial resolutions.

## 4. RESULTS

### 4.1 Results of quantitative comparison.

Under Wald's protocol setting, the GSA and MTF-GLP methods achieved the best performance in both images. GSA outperformed MTF-GLP on all metrics in the reconstruction of the 2022 hyperspectral cube, but MTF-GLP achieved better results for UIQI and SAM for the 2023 one (Table 1).

In the full-resolution evaluation (Table 2) HySure exhibited lower spectral distortion compared to the other two methods, albeit with a lower spatial quality, where MTF-GLP marginally outperformed the other methods.

**Table 1 - Wald's protocol assessment**

| Image | Method | UIQI | SAM | ERGAS |
|---|---|---|---|---|
| 2022-06-04 | GSA | **0.9809** | **6.7952** | **3.0607** |
| | MTF-GLP | 0.9806 | 6.8578 | 3.1245 |
| | HySure | 0.9764 | 7.0071 | 3.1051 |
| 2023-08-24 | GSA | 0.9698 | 7.1321 | **3.6788** |
| | MTF-GLP | **0.9710** | **7.1062** | 3.7721 |
| | HySure | 0.9596 | 7.7210 | 3.9407 |

**Table 2 - Metrics from full-resolution evaluation**

| Image | Method | $D_\lambda^k$ | $D_S^*$ | $Q^*$ |
|---|---|---|---|---|
| 2022-06-04 | GSA | 0.00844 | 0.55155 | 0.44466 |
| | MTF-GLP | 0.00850 | **0.55137** | **0.44482** |
| | HySure | **0.00811** | 0.55199 | 0.44437 |
| 2023-08-24 | GSA | 0.00982 | 0.51065 | 0.48454 |
| | MTF-GLP | 0.00988 | **0.51036** | **0.48480** |
| | HySure | **0.00875** | 0.51200 | 0.48372 |

### 4.2 Results of qualitative comparison.

Visual observation of RGB composite images (Fig.2) generated by the pansharpened images revealed that GSA and MTF-GLP approaches exhibit less noise and fewer colour alterations than HySure. However, the latter displayed higher contrast on both micro and global scales, effectively outlining features. MTF-GLP had a lower contrast level, concealing the most subtle features. The GSA-pansharpened images offered optimal feature visibility due to low noise and high contrast levels.

In s1 (Fig. 3), the pansharpening process improved the visibility of an embankment structure (blue arrow), adding relevant details useful for its accurate interpretation. Long traces of palaeochannels were visible in nearly every pansharpened image, owing to their substantial dimensions. PCs composite on GSA and HySure images enhanced bare soil trace contrast but couldn't highlight anomalies on vegetated fields. In contrast, MTF-GLP uniformly improved soil/vegetation trace visibility, albeit with reduced contrast.

In s2, a stretch of the Roman road known as Via Annia (blue arrow) with a NW-SE orientation was notably discernible on the GSA and HySure images. Similarly, the segment of a large fluvial ridge (yellow arrow) is noticeable in the SW part of the image. In s3, it is possible to discern an ancient artificial canal flowing towards SW, now buried (blue arrow). Additionally, several small palaeochannels are visible in its vicinity. The higher resolution of these images enables to correctly distinguish the narrow inner channels, ranging from 5 to 12 meters wide, from the broader shapes of the riverbanks. The GSA and HySure images distinctly highlight the trace of a palaeochannel in the eastern portion of the image (yellow arrow), while on the MTF-GLP image the trace is less pronounced.

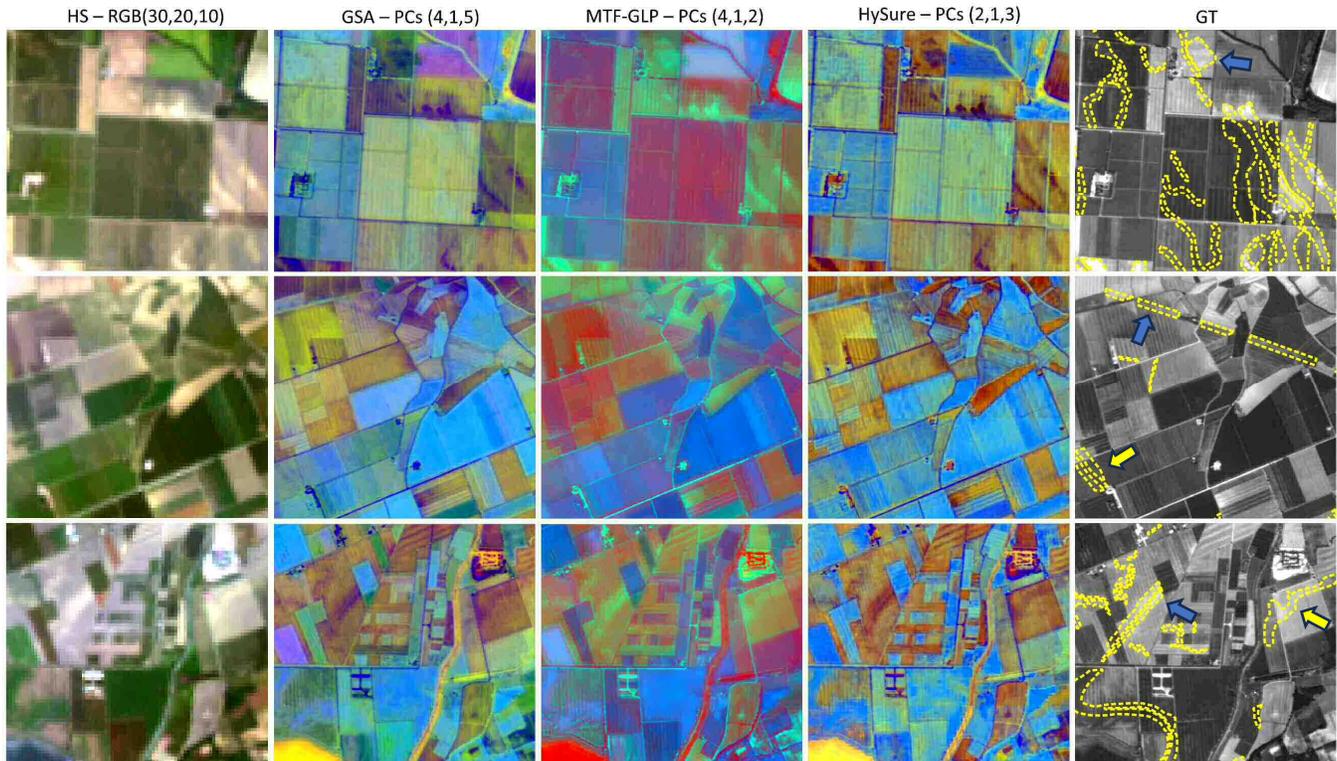

**Figure 3** – Comparison of HS product in RGB, false colours using PCs of GSA, MTF-GLP and HySure against geo-archaeological ground truth (GT) for s1, s2 and s3.

## 5. DISCUSSION

The significant improvement in the spatial resolution of PRISMA raw images resulting from pansharpening undoubtedly played a crucial role in the detection, and interpretation of archaeological and geomorphological traces. Specifically, for large-scale features such as palaeochannels, the enhanced resolution enables a more precise recognition of their boundaries. Pansharpening was also fundamental for visualising and interpreting small features like ancient (and buried) channels, structures and roads, characterised by dimensions smaller than those of a hyperspectral pixel. The application of PCA to the entire VNIR spectrum bands yielded marginal enhancements in making specific traces more visible. However, PCs are contingent on the context, and they cannot comprehensively improve the visibility of all anomalies across different parts of the territory.

In this study, we did not find a quantitative metric that serves as an accurate proxy for qualitative assessments in archaeological prospection. Surprisingly, MTF-GLP, despite achieving the best results in full-resolution evaluation, ranked lowest from a qualitative perspective. Nonetheless, a correlation was identified using Wald's protocol, albeit necessitating validation through further studies. Notably, for the June 2022 image, the GSA approach proved the most effective, establishing itself as the optimal choice not only in quantitative evaluation but also in terms of visual inspection.

## 6. CONCLUSION

This paper evaluated the applicability of pansharpened PRISMA's products to detect sub-surface archaeological features. The results underscore the effectiveness of pansharpening, establishing satellite hyperspectral data as a promising tool for advancing geo-archaeological prospection. Notably, a divergence is evident between quantitative and qualitative assessments, revealing a mismatch between the metrics used to rank pansharpening techniques and the practical utility of the obtained images.

This highlights the necessity for additional research in the development of metrics to automatically assess the visual potential of pansharpening techniques applied to hyperspectral data. Moreover, availability of higher resolution hyperspectral data from PRISMA will encourage the study of other geo-archaeological features and the exploitation of information in the SWIR portion of the spectrum, which is particularly efficient to detect features present in bare soil.

### Acknowledgement

This study was undertaken within the PERSEO project, co-funded by Agenzia Spaziale Italiana (ASI) in the framework of 'PRISMA Scienza' programme, grant agreement n. 2022-32-U.0.